\documentclass[accepted]{uai2021} 

\usepackage[utf8]{inputenc}
\usepackage[T1]{fontenc}
\usepackage{amssymb, amsmath,amsthm}
\usepackage[dvispcolors]{xcolor}
\usepackage{mystyle}
\usepackage{enumitem}
\usepackage[ruled,vlined]{algorithm2e}
\usepackage{pgfplots}
\usepgfplotslibrary{fillbetween}
\usepackage{caption}
\usepackage{subcaption}
\setlist[itemize]{leftmargin=*,nosep,nolistsep}

\pgfplotsset{compat=1.3}

\usepackage[american]{babel}

\usepackage{natbib} 
\usepackage{mathtools} 
\usepackage{booktabs} 
\usepackage{tikz}

\title{Improving Approximate Optimal Transport Distances using Quantization}

\author[1, 2]{\href{mailto:gaspardbeugnot@outlook.com}{Gaspard Beugnot \thanks{This work was conducted in large part during an internship at MIT.}}{}}
\author[1]{\href{mailto:aude.genevay@gmail.com}{Aude Genevay}{}}
\author[3]{\href{mailto:kristjan.h.greenewald@ibm.com}{Kristjan Greenwald}}
\author[1]{\href{mailto:jsolomon@mit.edu}{Justin Solomon}}

\affil[1]{%
    MIT CSAIL
}
\affil[2]{%
    INRIA
}
\affil[3]{%
    IBM Watson AI Lab
}

\begin{document}
\maketitle

\begin{abstract}
    Optimal transport (OT) is a popular tool in machine learning to compare probability measures geometrically, but it comes with substantial computational burden. Linear programming algorithms for computing OT distances scale cubically in the size of the input, making OT impractical in the large-sample regime. We introduce a practical algorithm, which relies on a quantization step, to estimate OT distances between measures given cheap sample access. We also provide a variant of our algorithm to improve the performance of approximate solvers, focusing on those for entropy-regularized transport. We give theoretical guarantees on the benefits of this quantization step and display experiments showing that it behaves well in practice, providing a practical approximation algorithm that can be used as a drop-in replacement for existing OT estimators.
\end{abstract}

Optimal transport (OT) is a versatile component of the probabilistic toolbox for machine learning.  As an alternative to conventional divergences between probability measures, OT provides a means of measuring how distributions align geometrically.  OT has found application in 
parameter estimation \citep{bernton2019parameter}, robust learning \citep{esfahani2018data}, and generative modeling \citep{salimans2018improving,genevay2017learning}---among other learning tasks.

When distributions are absolutely continuous or composed of huge numbers of points, it becomes infeasible to compute OT distances exactly.  In this setting, a common approximation follows two steps:  First, we draw $k$ samples from both distributions, and then we use linear programming to extract the distance between empirical distributions.  This plug-in procedure produces a convergent approximation as $k\to\infty$ (by the Glivenko--Cantelli theorem, since the Wasserstein distance metrizes weak convergence \citep{villani2003topics}), but two challenges conspire to limit its scalability:
\begin{itemize}
    \item \emph{Sample complexity} bounds and related results show that this approximation converges with rate $k^{-\nicefrac{1}{d}}$, where $d$ is the ambient dimension \citep{dudley1969speed,weed2017sharp}. These sharp \emph{asymptotic} rates exhibit a curse of dimensionality: we need a large number $k$ of samples (growing exponentially with $d$) before the approximation is useful.
    \item The \emph{computational complexity} of solving the linear program is roughly cubic in $k$ \citep{burkard2012assignment}, limiting the maximum $k$ we can take before this method becomes unreasonably slow.
\end{itemize}
Together, these facts imply that the largest $k$ for which solving the linear program is feasible may not be sufficient for extracting a usable distance estimate, i.e., the bottleneck is not availability of samples/data (the classic statistical setting), but \emph{computation} budget.

Our work is motivated by a simple observation about the methodology above. In machine learning, it is often straightforward to sample from the input measures for OT, e.g.\ when they come from large datasets, generative models, or easily-sampled smooth distributions. In this case, limited approximation quality is a byproduct of the cubic computational expense rather than a paucity of samples.  The algorithm above only draws $O(k)$ samples---but it could draw more without affecting the asymptotic runtime. That is, we can improve approximation quality with little added computational expense by drawing more than $k$ samples, cutting down to $k$ \emph{representative} (weighted) samples, and then solving a smaller discrete problem.

We introduce a practical, easily-implemented improvement to empirical OT. In our algorithm, the OT solver remains either the linear program solver or the recently-popular regularized Sinkhorn algorithm \citep{cuturi2013sinkhorn}. As input to this step, however, we ``summarize'' a superlinear number of samples with $k$ weighted samples through quantization. Our technique is seamless to implement given an implementation of empirical OT and substantially improves approximation quality given fixed computational cost. It can be used as a drop-in replacement for existing estimators. Beyond verifying performance empirically, we provide theory predicting the behavior we observe, in the low quantization error setting. While it is impossible to overcome the asymptotic curse of dimensionality associated to all finitely-supported measures \citep{kloeckner2012approximation}, our method leverages better convergence rates in the finite sample regime for ``clusterable'' distributions \citep{weed2017sharp}. This leads to substantial practical benefit, with an improvement of the exponent of the convergence rate by a factor $2$ in the best case (fast decaying tails) or at worst on par with the plug-in estimator (close to uniform). 

\textbf{Related work.}
OT suffers from a severe curse of dimensionality. Effective approximation requires an exponential number of samples $n$ in the ambient dimension. For an absolutely continuous measure $\mu$ (w.r.t.\ Lebesgue), its Wasserstein distance to any measure supported on $n$ points is asymptotically lower-bounded by $O(n^{-\nicefrac{1}{d}})$ \citep{dudley1969speed}.  This bound can sometimes be circumvented, e.g., when the measures have lower intrinsic dimension \citep{weed2017sharp} or when the support is discrete (convergence rate $O(\sqrt{\nicefrac1n})$, with constant depending on dimension) \citep{sommerfeld2018optimal}. To counter this curse of dimensionality, the best-known workaround relies on entropic regularization, with $O(\sqrt{\nicefrac1n})$ convergence \citep{genevay2018sample}. Another estimator penalizes the rank of the transport plan \citep{forrow2018statistical}, while \citep{goldfeld2020gaussian} proposes a smoothed distance by convolving measures with Gaussians. While these exhibit better convergence rates, they only approximate the Wasserstein distance and do not converge to its true value. 
The curse of dimensionality can sometimes be mitigated for standard OT---\citep{weed2017sharp} proves that for mixtures of Gaussians and clusterable distributions, the $p$-th power of the $p$-Wasserstein distance enjoys a $O(\sqrt{\nicefrac1n})$ rate for small $n$---implying a $O(n^{\nicefrac{-1}{4}})$ rate for $W_2$.

While the curse of dimensionality requires many samples to approximate transport reliably, in practice computational complexity prevents us from doing so. OT between discrete measures yields a large-scale linear program solvable using network flow solvers or the Hungarian algorithm, when both measures have the same size and uniform weights \citep{burkard2012assignment}. These take $O(n^3 \log n)$ time, where $n$ is the support size. As a faster alternative, entropy-regularized OT can be solved with quadratic complexity using Sinkhorn's algorithm \citep{sinkhorn1967diagonal}, but its convergence rate decays when regularization goes to zero \citep{franklin1989scaling}. 

For efficient OT approximation, we oversample the input measures and compute a summary via a quantization algorithm like $k$-means; note quantization is equivalent to finding the closest measure supported on $k$ points in 2-Wasserstein distance \citep{pollard1982quantization,canas2012learning}. The original $k$-means algorithm \citep{lloyd1982least} is prohibitive for large sample sizes and often reaches local minima. With a careful initialization, however, \citep{arthur2006k} proved that $k$-means likely converges to near its global optimum. This initialization, called $k$-means++, is obtained via $D^2$ sampling and is $O(\log k)$-close to optimal in expectation. This yields a cheap approximation in $O(nk)$ time, since the algorithm requires $k$ passes through the data. Later variants have lower computational complexity, among which \citep{bahmani2012scalable} performs only a fixed number of passes on the data and \citep{bachem2016fast} uses an MCMC $D^2$ sampler. These benefit from bounds similar to $k$-means++ but have $O(n)$ computational complexity.

Our approach has similarities with a line of work that uses a multi-scale scheme to compute optimal transport efficiently \citep{schmitzer2013hierarchical, gerber2017multiscale}. However, they focus on accelerating the exact computation of optimal transport, while we target a fast approximation. These multi-scale approaches also do not leverage a connection between $k$-means and optimal transport to yield quantitative analysis, and they are not applicable to entropy-regularized transport.

\paragraph{Contributions.} We propose efficient OT estimators using quantization, with theoretical analysis for two classes of OT problems:
\begin{itemize}
    \item \textbf{(Unregularized) OT:} We leverage the link between OT and $k$-means \citep{pollard1982quantization,canas2012learning} to quantify the bias and give precise bounds for Gaussian mixtures and clusterable distributions in the non-asymptotic regime.
    \item \textbf{Entropy-regularized OT}: Building on complexity results for Sinkhorn \citep{DBLP:journals/corr/AltschulerWR17}, we prove that our pre-processing can yield $\epsilon$-approximate OT with better time/space complexity.
\end{itemize}
We compare our estimators to the plug-in estimator on toy and real-world datasets. 

\textbf{Notation.}
Let $\mu$ and $\nu$ be probability measures on a compact set $\X\subseteq \RR^d$. The 2-Wasserstein distance between $\mu$ and $\nu$ is
\begin{equation}\label{eq:w2}
W_2(\mu,\nu)\!\eqdef\!\left( \min_{\pi \in\Pi(\mu,\nu) } \int_{\X\times\X} \hspace{-.2in} \|x-y\|^2_2\,d \pi(x,y)\right) ^{\nicefrac{1}{2}}\hspace{-.1in},
\end{equation}
where $\Pi(\mu,\nu)$ is the set of couplings on $\X\times\X$ with marginals $\mu,\nu$. 
Given $n$ samples from each measure, $X_n \eqdef(x_1, \dots, x_n) \sim \mu ^{\otimes n}$ and $Y_n \eqdef(y_1, \dots, y_n) \sim \nu ^{\otimes n}$, the \emph{empirical plug-in estimator} for $W_2$ is 
\begin{equation}\label{eq:plugin}
W_2(\hat \alpha_n,\hat \beta_n) =\!\left( \min_{\substack{\pi \ones = \ones/n \\ \pi^T \ones = \ones/n} } \frac{1}{n^2}\!\!\sum_{i,j=1}^n \|x_i-y_j\|_2^2 \pi_{ij}\right) ^{\nicefrac{1}{2}}\hspace{-.1in},
\end{equation}
where $\hat \alpha_n\eqdef \frac{1}{n}\sumin \delta_{x_i}$ and $\hat \beta_n\eqdef \frac{1}{n}\sumin \delta_{y_i}$ are empirical measures from $\mu$ and $\nu$, resp.

\section{Algorithm Overview}

We aim to improve the plug-in estimator $W_2(\hat \mu_k,\hat \nu_k)$, which approximates $W_2(\mu,\nu)$ with $O(k^3 \log k)$ computational complexity (that of LP solvers) and $O(k^{-\alpha})$ bias, given $k$ samples from each measure. In the worst case (e.g., uniform distributions), $\alpha = \nicefrac{1}{d}$, but there exist
regimes in which the rate improves (see \S\ref{sec:assumptions}). Our idea is to \emph{oversample} the measures, using $n > k$ samples to construct approximations of $\mu$ and $\nu$ of size $k$ that yield an estimated OT value with better bias
while preserving computational complexity. To satisfy these criteria, we need to ensure that pre-processing takes $O(k^3 \log k)$ time.

We denote by $\Shatk (X_n)$ a stochastic map that inputs a sample $X_n = (x_1, \dots, x_n) \sim \mu^{\otimes n}$ and outputs a $k$-point quantization. For any finite $S \subseteq \X$, use the function $P_S:\X \to S$ to denote the function that maps any point in $\X$ to its nearest neighbor in $S$. Denoting by $\hat \mu_n$ (resp., $\hat \nu_n$) the empirical measure associated to the $n$-sample $X_n$ (resp., $Y_n$) and $f_\#(\mu)$ the pushforward of $\mu$ through $f$, our estimator is defined as:
  \begin{equation}
    \label{our_estimator}
    \Est (k, n) \eqdef W_2(P_{\Shatk(X_n)\#}(\hat{\mu}_n), P_{\Shatk(Y_n)\#}(\hat{\nu}_n) ) .   
 \end{equation}
That is, we replace $\hat \mu_k$, $\hat \nu_k$ in the plug-in estimator \eqref{eq:plugin} with weighted $k$-point measures $P_{\Shatk(X_n)\#}(\hat{\mu}_n)$ and $P_{\Shatk(Y_n)\#}(\hat{\nu}_n)$, the centers of approximate $k$-means on $X_n$ and $Y_n$, resp. Each center is weighted proportionally to the number of samples in its Voronoi region. The plug-in estimator \eqref{eq:plugin} corresponds to $n = k$.

There are two steps in our pre-processing: \emph{(i)} selecting $k$ points representative of the larger $n$ samples and \emph{(ii)}  weighting the resulting $k$ points with the number of samples in their Voronoi regions. For $k$-means++, \emph{(i)} is $O(nk)$ while for \citep{bachem2016fast,bahmani2012scalable} it is $O(k)$. Regardless, the assignment in step \emph{(ii)} requires $O(nk)$ time. To be consistent with the $O(k^3 \log k)$ time complexity of the OT solver, we thus set $n = k^2 \log k$.

{
\begin{algorithm}[h]
  \caption{Approximation of $W_2(\mu, \nu)$}
  \label{our_algorithm}
\small
  \SetKwInOut{Input}{Input}
  \Input{Two samplers $\mu, \nu$; number of anchor points $k$.}
  
  \SetKwInOut{Output}{Output}
  \Output{Approximation of $W_2(\mu, \nu)$ with complexity $O(k^3 \log k)$}
  
  \tcc{Sample $n$ points}
  Set $n = k^2 \log k$ 
  
  Sample $X_n = (x_1, \dots, x_n)$ i.i.d.\ from $\mu$ and $Y_n = (y_1, \dots, y_n)$ i.i.d.\ from $\nu$
  
  \tcc{Subsample $k$ anchor points}
  Compute $\Shatk(X_n) = (c_1, \dots, c_k)$ with $k$-means++
  
  Compute $\Shatk(Y_n) = (d_1, \dots, d_k)$ with $k$-means++
  
  \tcc{Compute weights}
  Set $a_i = \sum_{j=1}^n \mathbf{1}_{i = \arg \min_{l} \lVert x_j - c_l \rVert^2_2}\ \forall i\in\{1,\ldots,k\}$  
  
  Set $b_i = \sum_{j=1}^n \mathbf{1}_{i = \arg \min_{l} \lVert x_j - d_l \rVert^2_2}\ \forall i\in\{1,\ldots,k\}$
  
  \tcc{Cost matrix}
  Set $\boldsymbol{C}_{ij} = \lVert c_i - d_j \rVert^2_2\ \forall i,j\in\{1,\ldots,n\}$
  
  \tcc{Weighted Wasserstein distance}
  \KwRet{$W_2(P_{\Shatk(X_n)_\#}\hat{\mu}_n, P_{\Shatk(Y_n)_\#}\hat{\nu}_n) \eqdef L_{\boldsymbol{C}}(a, b)^{\nicefrac12}$}
  
\end{algorithm}}

Algorithm \ref{our_algorithm} summarizes our estimator. It takes four steps: (1) sample $k^2 \log k$ points from each measure, (2) run $k$-means++ initialization, (3) project the $k^2 \log k$ points onto the $k$ cluster centers, and (4) compute OT between these new weighted point clouds. Steps (1) and (3) are seamless to implement, while steps (2) and (4) have readily available implementations in many languages, as they come from well-known algorithms. Thus, the procedure is highly practical, and it can easily be implemented to improve the bias of OT estimation with similar running times.

The performance of our approach is summarized informally in the theorem below; we bound bias 
in \S\ref{sec:theory}.
\begin{theorem}[Informal]
Algorithm \ref{our_algorithm} runs in $O(k^3\log k)$ time.\footnote{This complexity assumes sampling is cheap, i.e., $O(1)$. If drawing samples requires complex operations, the number of points we can sample will be below $n = k^2 \log k$; it is straightforward to adapt to this case.} The estimator has $O(k^{-2\alpha})$ bias in the best case and $O(k^{-\alpha})$ at worst, where the latter is the bias of the empirical plug-in estimator.
\end{theorem}

\begin{remark} The ``best case'' happens in the finite sample regime, when distributions have low quantization error as defined in \S\ref{sec:assumptions}. For near-uniform distributions, we get the asymptotic rate right away and cannot hope to improve on the plug-in estimator.
\end{remark}

This theorem predicts the performance observed in \S\ref{sec:experiments}. In short, with the same computational complexity, we improve the bias by an exponent of 2 compared to the plug-in estimator. Time complexity is a direct addition of pre-processing and LP solver complexities. The bias 
bounds, on the other hand, require more work and are the object of the next section.

\section{Theoretical analysis}\label{sec:theory}
\subsection{Bounding bias}
The bias of our estimator $\Est (k, n)$ defined in \eqref{our_estimator} is 
\begin{equation}
\label{main_objective}
    \Bias(k, n) = \absv{W_2(\mu, \nu) - \E \csb{\Est(k,n)}}.
\end{equation}
By the triangle inequality on $\absv{\cdot}$ and $W_2$, we have that 
\begin{align*}
\Bias(k, n) \leq  &\E_{X_n, \Shatk} \csb{W_2(\mu, P_{\Shatk(X_n)\#}(\hat{\mu}_n))} \\ 
+ &\E_{Y_n, \Shatk} \csb{W_2(\nu, P_{\Shatk(Y_n)\#}(\hat{\nu}_n))},
\end{align*}
so bounding bias amounts to controlling the two terms above.
This requires some definitions:
\begin{definition}[Quantization error $\phi_S(C)$]
\label{bachem_kmeans_notation}
Let $\C \subseteq \X $ be a finite set of $n$ elements. For any $S \subseteq \X$, define the \emph{quantization error} of $\C$ w.r.t.\ $S$ as 
    \[\phi_S(\C) = \sum_{x \in \C} d(x, S)^2,\] where 
    $d(x, S) = \min_{s\in S} d(x,s)$. 
For $k \leq n$, denote by $\kmeansOPT$ the \emph{optimal quantization error} for a set of $k$ elements, written
    $\kmeansOPT(\C) = \min_{S \subseteq \X, \absv{S} = k} \phi_S (\C)$, 
and $S_k$ its minimizer. 
\end{definition}

We can relate the bias of our estimator to sample complexity and quantization error as follows:
\begin{theorem}[Bias of the estimator]
\label{thm:optimal_bias_estimator}
Suppose 
    $\E\csb{W_2(\mu, \hat \mu_n)} \leq O(n^{-\alpha}),$ 
where $\alpha$ is the sample complexity rate of $\mu$. Then, for a sample $X_n \sim \mu^{\otimes n}$, 
\begin{align*}
    \E_{X_n, \Shatk} &\csb{W_2(\mu, P_{\hat S_k(X_n)\#} \hat{\mu}_n)} \leq  \\&\qquad O\p{n^{-\alpha} + \sqrt{\nicefrac{(\log k)}{n}}\E_{X_n} \csb{\kmeansOPT(X_n)}^{\nicefrac{1}{2}}}.
\end{align*}
\end{theorem}
The sample complexity here is not necessarily the asymptotic rate $\alpha=\nicefrac{1}{d}$. Rather, we will see in \S\ref{sec:assumptions} that our estimator performs well in the finite sample regime for clusterable distributions, with rate $\alpha=\nicefrac{1}{4}$.
\begin{proof}
By the triangle inequality on $W_2$, we can decompose into two quantities $A$ and $B$:
\begin{align}
    \E_{X_n, \Shatk} \csb{W_2(\mu, P_{\hat S_k(X_n)\#} \hat{\mu}_n)} \leq & \nonumber \\
    \underbrace{\E_{X_n} W_2(\mu, \hat{\mu}_n)}_{A}
     + 
    \underbrace{\E_{X_n, \Shatk} W_2(\hat{\mu}_n, P_{\hat S_k(Xn)\#} \hat{\mu}_n)}_{B} \label{n_k_decomposition}
.
\end{align}
\vspace{-5mm}
\begin{itemize}
    \item $A$ is the sample complexity rate of the empirical distribution, which we assume to be $O(n^{-\alpha})$. 
    \item $B$ is the error made when projecting the $n$ samples onto $k$ weighted points chosen by {$k$-means++}. If $n=k$, it vanishes and we recover the sample complexity of the empirical estimator. 
    Controlling $B$ requires relating Wasserstein distance to the optimal quantization \citep{canas2012learning}.
\end{itemize}
Denoting $X_n = (x_1, \dots, x_n) \sim \mu ^{\otimes n}$, 
we write: 
 \begin{align}
 B &= \E_{X_n, \Shatk} \csb{\p{\frac{1}{n} \sum_{i=1}^n d(x_i, \hat S_k)^2 }^{\nicefrac{1}{2}}}\nonumber\\
 &\leq \E_{X_n} \csb{\p{
     \frac{8(\log k + 2)}{n} \kmeansOPT(X_n)
     }^{\nicefrac{1}{2}}}.\label{eq:B}
 \end{align}
The first equality comes from the equivalence between $W_2$ and the quantization error \cite[Lemma 1]{canas2012learning}. The second is 
the $k$-means++ optimality bound of \citep{arthur2006k}. Jensen's inequality completes the proof. Note having the optimal set $S_k$ instead of  $\hat S_k$ would remove the $\log k$ factor in \eqref{eq:B}.
\end{proof}

In our algorithm, we take $n=k^2 \log k$ to get the following bias for our estimator:
\begin{corollary}In the setting of Theorem~\ref{thm:optimal_bias_estimator}, with $n=k^2 \log k$, our estimator (Algorithm~\ref{our_algorithm}) satisfies
 \begin{align*}
    \E_{X_n, \Shatk} &\csb{W_2(\mu, P_{\Shatk(X_n)\#}(\hat{\mu}_n))} \leq \\&\qquad O\p{(k^2 \log k)^{-\alpha} + \nicefrac{1}{k}\E_{X_n} \csb{\kmeansOPT(X_n)}^{\nicefrac{1}{2}}}.
 \end{align*}
\vspace{-3mm}
\label{cor:bound}
\end{corollary}
Corollary~\ref{cor:bound} tells us that at best, our estimator improves the exponent in the bias bound by a factor of 2, going from $O(k^{-\alpha})$ to $O(k^{-2\alpha})$ while keeping computational complexity on par with the empirical plug-in estimator. To benefit from this improvement, we need to ensure quantization error---the second term in the bound---is small enough so that the first dominates.

\subsection{Controlling the quantization error}\label{sec:assumptions}

To prove our estimator improves bias, we must make an assumption on the behavior of the quantization error when quantizing an $n$-sample from $\mu$ on $k$ points. 
Intuitively, the quantization error is small when the measure is well-concentrated. In particular, we can upper bound quantization error for Gaussian mixtures and measures supported on finite numbers of balls.
 
 \begin{remark} We derive improved theoretical rates for these two classes of functions, but our algorithm is better than the plug-in estimator for \emph{any} dataset whose quantization error is smaller than the sample complexity.  This is verified by several real-world datasets (Fig.\ref{fig:qerr}, supplement), underscoring the practical significance of our proposed algoritm.
 \end{remark}
\begin{definition}[Clusterable distribution]
A distribution $\mu$ is an $(m,\sigma^2)$-Gaussian mixture if it is a mixture of $m$ Gaussian distributions in $\RR^d$ and the trace of the
covariance matrix of each mixture component is upper-bounded by $\sigma^2$. 
A distribution $\mu$ is $(m,\Delta)$-clusterable if $\mathrm{supp}(\mu)$ lies in the union of $m$ balls of radius at most $\Delta$.
\end{definition}
By writing down the definition of $\phi_{k}^{OPT}$, it is straightforward to prove that for k $\geq m$, $\nicefrac{1}{n} \cdot \EE[\phi_{k}^{OPT}({X_n})] \leq \sigma^2$ if $\mu$ is a $(m,\sigma^2)$-Gaussian mixture, and $\nicefrac{1}{n} \cdot \EE[\phi_{k}^{OPT}({X_n})] \leq \Delta^2$ if $\mu$ is $(m,\Delta)$-clusterable.

Incidentally, for such measures, better sample complexity rates can be derived \citep{weed2017sharp}:
\begin{proposition}[\citep{weed2017sharp}]\label{prop:weed_bach_finite_sample}
If $\mu$ is a $(m,\sigma^2)$-Gaussian mixture and $\log \frac{1}{\sigma}\geq 25/8$, then for all $n \leq m (32 \sigma^2 \log \frac{1}{\sigma})^{-2}$,
 \begin{equation}\label{eq:weedbach}
 \mathbb{E}[W_2^2(\mu, \hat{\mu}_n)] \leq 84 \sqrt{\nicefrac{m}{n}}.
\end{equation}
  The same rate holds for $(m,\Delta)$-clusterable distributions, for all $n \leq m (2\Delta)^{-4}$.
\end{proposition}

This result can be extended to distributions that are mixtures with fast decaying tails. This improved rate holds \emph{in the small-sample regime}, but asymptotically, the $\nicefrac1d$ rate returns. This rate is for \emph{squared} $W_2$, so in our analysis using $W_2$ this only implies $\alpha = \nicefrac14$ via Jensen's inequality. Thus, these improved rates for $W_2$ are only relevant in dimension higher than 4.

Further assumptions on $\sigma^2$ (resp.\ $\Delta$) improve the convergence rate of the bias from Theorem~\ref{thm:optimal_bias_estimator}:
\begin{proposition} \label{prop:cluster_rate} If $\mu$ is an $(m,\sigma^2)$-Gaussian mixture (resp.\ $(m,\Delta)$-clusterable), then for all $k \geq m$ such that $k^2\log k  \leq m (32 \sigma^2 \log \frac{1}{\sigma})^{-2}$ (resp.\ $k^2\log k \leq m (2\Delta)^{-4}$) our estimator (Algorithm \ref{our_algorithm}) satisfies 
\begin{align*}
  \mathbb{E}[W_2(\mu, P_{\hat S_k(X_n)\#}\hat{\mu}_n)]  \leq &\sqrt{84}\left( \frac{m}{k^2\log k}  \right)^{\!\!\nicefrac14}\!\!\!\!+\!C \sigma \sqrt{\log k},
\end{align*}
(replacing $\sigma$ by $\Delta$ in the above bound for clusterable distributions), where $C$ is  
independent of $k$ and $\sigma$. 
 If 
 $k^2\log k  \leq m (32 \sigma^2 \log \frac{1}{\sigma})^{-2}$ (resp. $k^2\log k \leq m (2\Delta)^{-4}$),
 then
 $\sigma \leq O((\log k)^{\nicefrac{-1}{4}}\:k^{\nicefrac{-1}{2}})$ (resp. $\Delta$), and
 the rate becomes $O((\log k)^{\nicefrac{1}{4}}\:k^{\nicefrac{-1}{2}})$.
 \end{proposition}
 
 Hence, we achieve an $O((\log k)^{\nicefrac{1}{4}}\:k^{\nicefrac{-1}{2}})$ rate in $O(k^3)$ computation time, compared to the $O(k^{-\nicefrac14})$ rate of the empirical estimator. 
 For the range of $k$ we consider, we observe in practice that the assumption $\nicefrac{1}{n}\cdot \phi_{k}^{OPT}\leq \nicefrac{1}{k}$ often holds, and hence our bound applies. Due to the curse of dimensionality, however, there is no guarantee for this to hold in the asymptotic case. 
 
 \paragraph{Intuition on the finite sample regime.} 
 The intuition for the bound of Proposition \ref{prop:weed_bach_finite_sample} is not simple. We provide an informal explanation.
 From a high level, in the small sample regime, we are looking at a coarse scale (e.g. from a distance, Gaussians ``look like'' Diracs) so the bound behaves like discrete optimal transport, which is $n^{-1/2}$. However when the number of samples grows, we are looking at a fine scale; in this regime, we suffer from the curse of dimensionality. A second piece of intuition is simpler: when you have very few samples, every new sample brings a lot of information, but after a while, the information gain of each new sample diminishes.

\section{Regularized Transport}\label{sec:regularized_transport}

Quantization can also improve approximate OT solvers, as it introduces negligible error while improving the required runtime and memory storage, at least in the discrete case.
We focus on entropic regularization, a popular approximation of OT obtainable in quadratic time with Sinkhorn's algorithm \citep{cuturi2013sinkhorn}. 
More precisely, the computational complexity to obtain an $\epsilon$-approximation of the unregularized cost for discrete problems is bounded by $O(k^2\epsilon^{-2})$, an order of magnitude cheaper than the linear program \citep{sinkhornbound}. 
The oversampling strategy used previously for absolutely continuous measures is irrelevant, however: quantizing $n$ points with $k$ centroids takes at least $O(nk)$ time (because of weight assignment), which exceeds $O(k^2)$ for $n > k$. 

Instead, we consider the case where we are given two very large discrete measures as input and rely on quantization to design a more efficient approximation procedure. In this setting, the literature focuses on \textit{complexity bounds}: given two discrete distributions over $n$ points and a target precision $\epsilon$, the aim is to provide an $\epsilon$-approximation of unregularized transport with bounded complexity \citep{DBLP:journals/corr/AltschulerWR17, DBLP:journals/corr/abs-1802-04367, sinkhornbound}. Building on this problem formulation, we propose a quantization step with target precision $\epsilon$ as a preprocessing step. Afterwards, any approximate transport solver can be used on the resulting quantized distribution. This provides the same theoretical guarantees and bounded computational complexity as above, with
potential computation time improvements. Our algorithm is detailed in Algorithm \ref{our_algorithm_for_regularized_transport}.

{
\begin{algorithm}[h]
  \caption{$\epsilon$-approximation of $W_2(\mu_n, \nu_n)$}
  \label{our_algorithm_for_regularized_transport}

\small
  \SetKwInOut{Input}{Input}
  \Input{Finite distributions $\mu_n, \nu_n$; target precision $\epsilon$}
  
  \SetKwInOut{Output}{Output}
  \Output{$3 \epsilon$-approximation of $W_2(\mu_n, \nu_n)$ with complexity $O(k^2 \epsilon^{-2})$}
  
  \tcc{Quantize the point clouds}
  $S_\epsilon = \quantize\,(\mu_n, \epsilon)$; $\absv{S_\epsilon} = k_{\epsilon, \mu_n}$
  
  $T_\epsilon = \quantize\,(\nu_n, \epsilon)$; $\absv{T_\epsilon} = k_{\epsilon, \nu_n}$
  
  \tcc{Compute weights and cost matrix}
  Set $a_i\!=\!\sum_{j=1}^n\!w_{\mu, j} \mathbf{1}_{i = \arg \min_{l} \lVert x_j - c_l \rVert^2_2}\ \forall i\!\in\!\{1,\ldots\!,k_{\epsilon, \mu_n}\}$  
  
  Set $b_i\!=\!\sum_{j=1}^n\!w_{\nu, j} \mathbf{1}_{i = \arg \min_{l} \lVert y_j - d_l \rVert^2_2}\ \forall i\!\in\!\{1,\ldots\!,k_{\epsilon, \nu_n}\}$
  
  Set $\boldsymbol{C}_{ij} = \lVert c_i - d_j \rVert^2_2\ \forall c_i,d_j\in S_\epsilon \times T_\epsilon$
  
  \tcc{Regularized transport solver}
  \KwRet{$\approxsolver \,(\boldsymbol{C}, a, b, \epsilon)$}
\end{algorithm}}

Algorithm \ref{our_algorithm_for_regularized_transport} relies on two subroutines: $\quantize$ and $\approxsolver$. The former inputs a point cloud $\mu_n$ and a tolerance $\epsilon$ and outputs a (sub)set $S_\epsilon$, which is a quantized version of $\mu_n$. \kmeanspp\ can be adapted easily to do this. An example is in Algorithm \ref{alg:kmeanspp_fixed_precision}. 
$\approxsolver$ yields an $\epsilon$ approximation of unregularized transport. The most used one is probably the Sinkhorn algorithm, which has a complexity bounded by $O(k^2\epsilon^{-2})$; see \citep{DBLP:journals/corr/AltschulerWR17} for details. This is the one we use in our experiments.  

{
\begin{algorithm}[h]
  \caption{$\quantize$}
  \label{alg:kmeanspp_fixed_precision}
\small
  \SetKwInOut{Input}{Input}
  \Input{A finite distribution $\mu_n$ with support and weights $(x_i, w_i)_{1\leq i \leq n)}$; target precision $\epsilon$.}
  
  \SetKwInOut{Output}{Output}
  \Output{Set $S_\epsilon$ with $k_\epsilon$ elements, s.t.\ $W_2^2 (\mu_n, P_{\hat{S}_\epsilon\#} \mu_n) = \sum_i w_i d(x_i, S_\epsilon)^2 < \epsilon^2$.}
  $S_\epsilon \leftarrow x_{\textsc{RAND}(1, n)}$
  
  $D = (w_i d(x_i, S_\epsilon)^2)_{1 \leq i \leq n}$

  \While{$\sum_i D_i > \epsilon^2$}{
    $S_\epsilon \leftarrow x_{\arg \max_i D_i}$
    
    $D = (w_i d(x_i, S_\epsilon)^2)_{1 \leq i \leq n}$
  }
  \KwRet{$S_\epsilon$}
\end{algorithm}}

Algorithm \ref{alg:kmeanspp_fixed_precision} is 
directly adapted from the original \kmeanspp\  algorithm. It is guaranteed to finish, as $S_\epsilon = \mu_n$ is a solution for any $\epsilon$. Denoting $k_\epsilon = \absv{S_\epsilon} \leq n$, we have that the complexity of Algorithm \ref{alg:kmeanspp_fixed_precision} is bounded by $O(n k_\epsilon)$. Thus, Algorithm \ref{our_algorithm_for_regularized_transport} has a complexity bounded by $O(n k_\epsilon + k_\epsilon^2 \epsilon^{-2}) \lesssim O(n^2\epsilon^{-2})$. The fact that it outputs a $3\epsilon$ approximation of OT relies on Lemma 1 of \citep{canas2012learning}:
\begin{equation}
\begin{aligned}
    W_2(\mu, \nu) \leq &W_2 (P_{\hat{S}_\epsilon\#} \mu,P_{\hat{T}_\epsilon\#} \nu)
    \\&+ W_2 (\mu, P_{\hat{S}_\epsilon\#} \mu) + W_2 (\nu, P_{\hat{S}_\epsilon\#} \nu)
\end{aligned}
\end{equation}
The first term is approximated within $\epsilon$ thanks to $\approxsolver$, the second/third 
thanks to Algorithm \ref{alg:kmeanspp_fixed_precision}.

Overall, we have two options to obtain a $3\epsilon$ approximation of $W_2(\mu_n, \nu_n)$:
\begin{itemize}
    \item Run $\approxsolver(\boldsymbol{C}_n, w_\mu, w_\nu, 3\epsilon)$ where $\boldsymbol{C}_n$ is the $n \times n$ cost matrix between $\mu_n$ and $\nu_n$.
    \item Run Algorithm \ref{our_algorithm_for_regularized_transport}. 
\end{itemize}
Both have a complexity $\leq O(n^2\epsilon^{-2})$ and provide the same theoretical guarantees; but the latter can provide a significant speed up. We compare both approaches in the next section, measuring CPU-time vs.\ precision.

\textbf{Space complexity.} While Sinkhorn's algorithm has space complexity of $O(n^2)$, we highlight that alg.\ \ref{our_algorithm_for_regularized_transport} has space complexity of $O(n + k_\epsilon^2)$. Indeed, the $\quantize$ algorithm only needs to keep track of the assignment of every point to their nearest centroid: this is a vector of size $n$. Thus, for huge datasets where storage is critical, quantization is a natural way to downscale the point cloud while keeping track of the precision loss. 

\begin{remark}
Some remarks about Algorithm \ref{our_algorithm_for_regularized_transport}:
\begin{itemize}
    \item The bound on the complexity of $\approxsolver$ usually involves $\norm{C}_\infty$. It will be smaller for the cost between centroids, providing additional speedup.
    \item This preprocessing step can be used for any $p$-Wasserstein distance, by changing the exponent in $\quantize$ accordingly ($D=(w_i d(x_i, S_\epsilon)^p)_{1 \leq i \leq n}$).
    \item We provide an algorithm with the same approximation guarantees than the baseline, with lower or equal computational complexity. 
    A sharp bound on the output of algorithm \ref{our_algorithm_for_regularized_transport} would require studying $\epsilon \mapsto k_\epsilon$. 
\end{itemize}
\end{remark}

\section{Experiments}\label{sec:experiments}

\textbf{Datasets.} We test on discrete (mainly real-world data) and continuous (synthetic) distributions. The latter tests theoretical bounds, while the former shows efficiency of Algorithm \ref{our_algorithm} on large point clouds. 
Fig.\ 5
(supplement) shows examples. The discrete datasets are: \emph{DOT}, \emph{Adult}, and \emph{Sampled Mixtures}. The `true' distance is computed on the whole point cloud; some datasets were downsampled to suit ground truth computation on our machine.
\emph{DOT} \citep{Schrieber_2017_DOT} contains grayscale images (i.e., fixed discrete support in $\RR^2$) in various resolutions, a benchmark used e.g.\ in \citep{sommerfeld2018optimal}, which uses the plug-in estimator. \emph{Adult} (UCI repository) is a point cloud in $\RR^6$ with continuous features for 35,000 individuals, split into two groups by income. \emph{Sampled Mixtures} (synthetic) contains 10,000 points from a Gaussian mixture with covariance $\tau$ in $\RR^{15}$, simulating point clouds suited to $k$-means. The continuous distributions are  \emph{Gaussians} and \emph{fragmented-hypercube} \citep{forrow2018statistical},\footnote{What they refer to as ``k-means \& OT'' is \emph{not} our Algorithm \ref{our_algorithm}, since they set $k=4$. Their $x$-axis does not relate to overall computational complexity.} with closed-form $W_2$; see Appendix 1
for details and more experiments. 

\subsection{Algorithm 1}
\begin{figure*}
     \centering
     \begin{subfigure}[t]{1.65in}
         \centering
         \includegraphics[width=\textwidth]{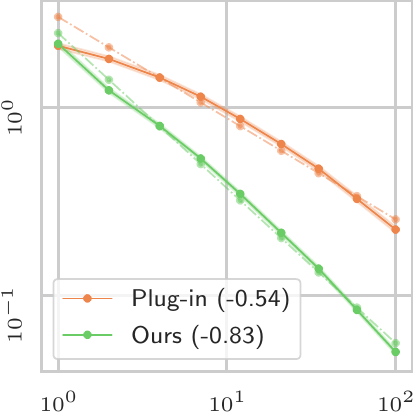}
         \caption{DOT dataset.}
         \label{result_DOT}
     \end{subfigure}
     \hfill
     \begin{subfigure}[t]{1.65in}
         \centering
         \includegraphics[width=\textwidth]{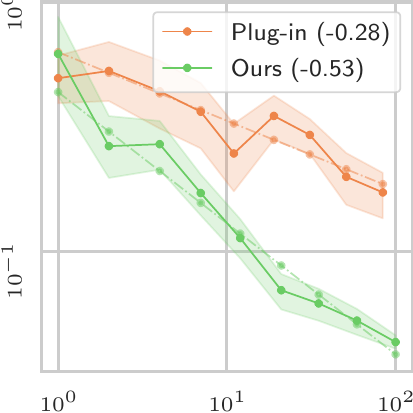}
         \caption{Adult dataset.}
         \label{result_adult}
     \end{subfigure}
    \hfill
     \begin{subfigure}[t]{1.65in}
         \centering
         \includegraphics[width=\textwidth]{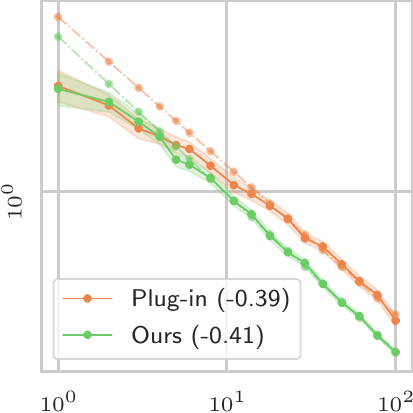}
         \caption{Discrete mixture, large variance ($\tau=0.1$).}
          \label{result_spreadout}
     \end{subfigure}
     \hfill
          \begin{subfigure}[t]{1.65in}
         \centering
         \includegraphics[width=\textwidth]{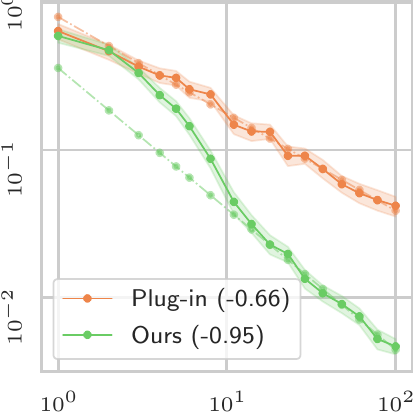}
         \caption{Discrete mixture, low variance ($\tau=10^{-4}$).}
          \label{result_synthetic_compact}
     \end{subfigure}
\vspace{-8pt}
\caption{Mean relative error vs.\ $k$ on discrete datasets. Values in parentheses display the regression coefficient computed for the second half of the graph. In \emph{(a)}, we plot the
average value of the 45 pairwise estimation on the DOT dataset ("Microscopy" images, 64 resolution).}
\label{fig:discrete_results}
\vspace{-12pt}
\end{figure*}

\begin{figure*}
     \centering
     \begin{subfigure}[t]{1.65in}
         \centering
         \includegraphics[width=\textwidth]{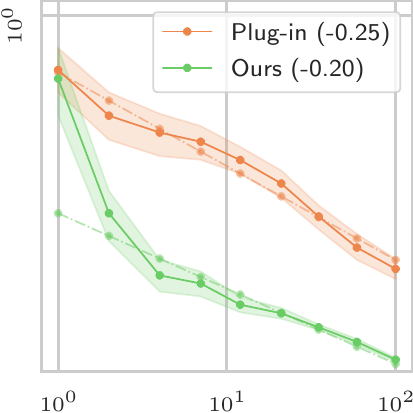}
         \caption{Gaussian with $10^{-1}$ diagonal covariance.}
         \label{fig:gaussian_0}
     \end{subfigure}
     \hfill
     \begin{subfigure}[t]{1.65in}
         \centering
         \includegraphics[width=\textwidth]{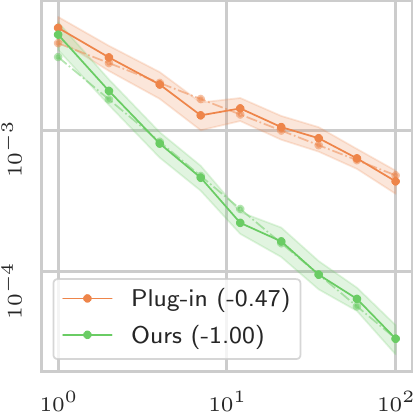}
         \caption{Gaussian with $10^{-4}$ diagonal covariance.}
         \label{fig:gaussian_4}
     \end{subfigure}
    \hfill
     \begin{subfigure}[t]{1.65in}
         \centering
         \includegraphics[width=\textwidth]{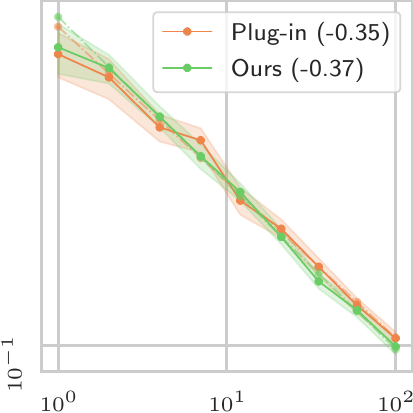}
         \caption{Fragmented hypercube, $d=8$.}
          \label{fig:hypercube_dim_8}
     \end{subfigure}
     \hfill
          \begin{subfigure}[t]{1.65in}
         \centering
         \includegraphics[width=\textwidth]{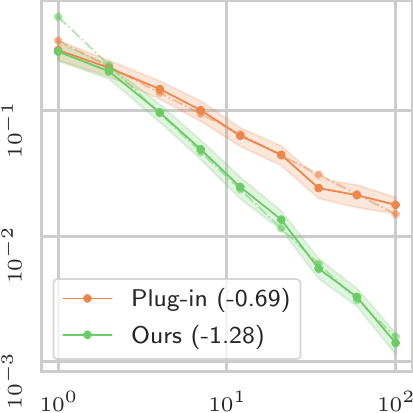}
         \caption{Fragmented hypercube, $d=2$.}
          \label{fig:hypercube_dim_2}
     \end{subfigure}
\vspace{-8pt}
\caption{Mean relative error vs.\ $k$ on continuous distributions. Values in parentheses display the regression coefficient computed for the second half of the graph. Left: Gaussian in $\RR^5$. When the clusterable assumption does not hold, the improvement is negligible. However, when the finite sample rate is applicable, the improvement is striking ($\times 2.1$). Right: Fragmented hypercube \citep{forrow2018statistical}. In high dimension, it resembles the uniform distribution and we get no improvement. In small dimension, the improvement is significant ($\times 1.8$).}
\label{fig:continuous_results}
\vspace{-12pt}
\end{figure*}

\begin{figure*}
\begin{minipage}[t]{\linewidth}
     \centering
     \begin{subfigure}[t]{2.22in}
         \centering
         \includegraphics[width=\textwidth]{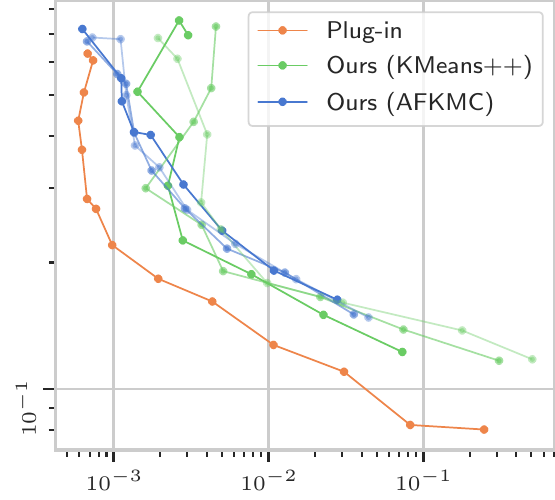}
         \caption{Spread Gaussian}
     \end{subfigure}
     \hfill
     \begin{subfigure}[t]{2.22in}
         \centering
         \includegraphics[width=\textwidth]{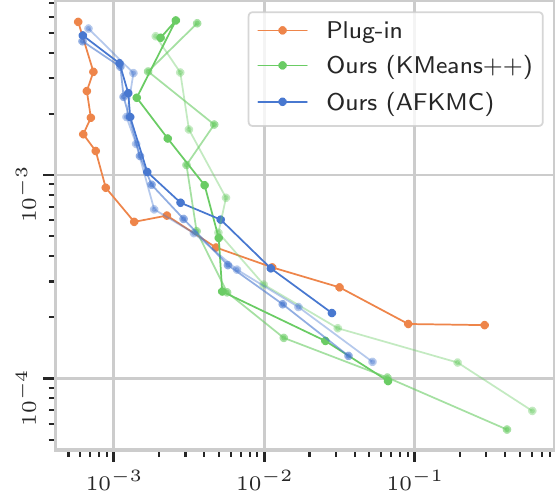}
         \caption{Peaked Gaussian}
     \end{subfigure}
    \hfill
     \begin{subfigure}[t]{2.22in}
         \centering
        \includegraphics[width=\textwidth]{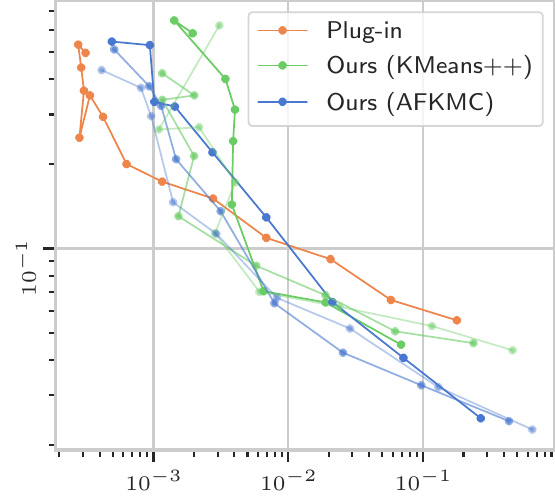}
         \caption{Adult Dataset}
     \end{subfigure}
\vspace{-8pt}
\caption{Mean relative error vs.\ CPU time ($s$) on (a) Gaussians with unit covariance, (b) Gaussians with $10^{-4}$ diagonal covariance, (c) Adult dataset. One line corresponds to log-spaced values of $\boldsymbol{k}$. Line's transparency correspond to various $\boldsymbol{\kappa} \in \cb{1, 0.5, 0.1}$, darkest for biggest value. Line's color corresponds to various \textbf{estimator}. We compare the plug-in estimator (orange) to two variants of our algorithm : $k$-means++ (green) or AFK-MC$^2$ (blue) from \citep{bachem2016fast} as a preprocessing step. For data with small quantization error, our approximate $k$-means pre-processing (even unoptimized) provides a clear advantage. }
\label{fig:cpu_time}
\vspace{-8pt}
\end{minipage}
\end{figure*}

For each dataset, we compare the behavior of the plug-in estimator and that of Algorithm \ref{our_algorithm}. We plot the mean \textit{relative} error $\E_{X_n, \Shatk} \csb{|\Est (k, k^2 \log k) - W_2 (\mu, \nu)|} / W_2 (\mu, \nu)$, estimating the expectation with 100 runs. We display two types of plots: \emph{(i)} mean relative error vs.\ $k$ (size of the point clouds passed to the LP) (Figures~\ref{fig:discrete_results},~\ref{fig:continuous_results}) and  \emph{(ii)} mean relative error vs.\ CPU time (Figure~\ref{fig:cpu_time}).

\textbf{Results.}
Our estimator exhibits favorable behavior when estimating $W_2$ between large \emph{point clouds}. In this case, the sample complexity of the plug-in estimator $W_2(\mu, \hat{\mu}_k)$ decays in $O(k^{-\nicefrac{1}{2}})$, independently of the dimension or number of samples (these only affect the constant  \citep{sommerfeld2018optimal}), but ours enjoys a faster decay rate exponent---up to twice better. For continuous distributions, our results are similarly advantageous in the finite-sample regime for clusterable distributions but tend to the sample complexity rate in higher dimensions. They provide a way to verify Theorem \ref{thm:optimal_bias_estimator} and to illustrate the different regimes. We notice in practice that oversampling enables the estimator to have much lower variance (fig. 7, supplement). 

\textbf{Discrete datasets.} On the real-world datasets, the bias decays 45\% (\emph{DOT}, fig.\ \ref{result_DOT}) to 65\% (\emph{Adult}, fig.\ \ref{result_adult}) faster. A simple analysis explains this: On a $100\!\times\!100$ image, with $k\leq\!100$ samples the plug-in estimator will sample $\sim\!1\%$ of the image, whereas our estimator processes all the pixels and then subsamples the 100 most relevant. Synthetic experiments slightly qualify this analysis: When the data is well-clustered 
the improvement is up to twice the decay rate (fig.\ \ref{result_synthetic_compact}), as expected from Proposition~\ref{prop:cluster_rate}; however, when the point cloud is more spread out, the decay rate only marginally improves over plug-in estimation.

\textbf{Continuous distributions.}
The plug-in estimator on Gaussian data recovers the expected $\nicefrac{-1}d$ rate exponent when variance is high (fig.\ \ref{fig:gaussian_0}); when the variance is low, we find the better finite sample complexity rate of $\nicefrac{-1}{2}$ predicted by \citep{weed2017sharp}. In this regime, our estimator beats the plug-in estimator by a large margin (fig.\ \ref{fig:gaussian_4}). Asymptotically, both curves should reach the same slope of $\nicefrac{-1}{d}$. 
Similarly, we should expect our estimator to degrade on the uniform distribution: for uniformly-spread data, quantization error decays in $k^{-\nicefrac{1}{d}}$. The \emph{Fragmented Hypercube} example confirms this: When $d = 2$, the distribution is clusterable (fig.\ \ref{fig:hypercube_dim_2}), but as $d$ increases
the quantization error is relatively high, eventually reaching the performance of the plug-in estimator (fig.\ \ref{fig:hypercube_dim_8}). 

\textbf{CPU time.}
Since our goal is to provide a faster $W_2$ approximation, we check the decay of the bias against CPU time. These experiments evaluate to what extent the theoretical improvement of the bias may be cancelled by overhead in $k$-means computation. The solver we use for OT \citep{flamary2017pot} is thoroughly optimized, making the comparison difficult. However, our estimator is only slower by a constant on spread out data (Figure~\ref{fig:cpu_time}(a)) and provides a clear advantage on clustered (Figure~\ref{fig:cpu_time}(b)) and real data  (Figure~\ref{fig:cpu_time}(c)). To further improve, \emph{(i)} our basic implementation of $k$-means++ could be optimized and \emph{(ii)} we can use theoretically weaker minimizers of the quantization problem. In Figure~\ref{fig:cpu_time}, we use a faster approximate quantizer, AFK-MC$^2$ \citep{bachem2016fast} with fixed chain length on $n=k^2 \log k$ points  (blue), which has overall complexity $k^2 \log k$ but weaker guarantees on the quantization error. Another alternative is to multiply the number of points used to compute the anchors (we tested $\kappa \in \cb{1, 0.5, 0.1}$) to further decrease the complexity constant between the pre-processing and the OT estimation steps. This can be used as a hyper-parameter to balance faster execution with lower bias improvement. For these experiments, we use an Intel(R) Core(TM) i5-7200U CPU @ 2.50GHz processor, with 8 GB memory. The $k$-means and OT solvers are implemented in C and wrapped in Python.

\begin{figure}[h]
\centering
    \begin{subfigure}[t]{1.6in}
         \includegraphics[width=1.6in]{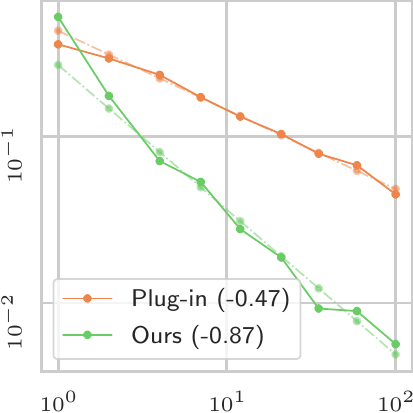} 
         \caption{Gaussian with unit diagonal covariance}
    \end{subfigure}
    \begin{subfigure}[t]{1.6in}
         \includegraphics[width=1.6in]{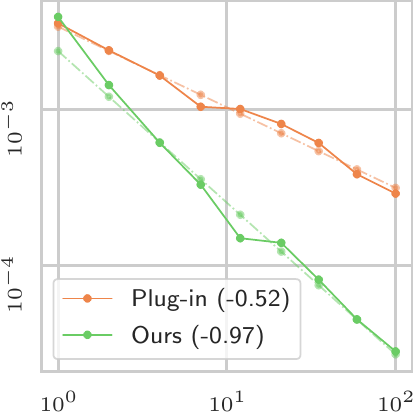} 
         \caption{Gaussian with $10^{-4}$ diagonal covariance}
    \end{subfigure}
\caption{Empirical standard deviation of Algorithm 1 vs. $k$. Sampling $k^2 \log k$ samples instead of $k$, our estimator manages a much lower standard deviation, independently of the clusterability of the distribution.}
\label{fig:standard_deviation_gaussian_alg1}
\end{figure}

\paragraph{Variance of the estimator.} Algorithm \ref{our_algorithm} relies on \textit{oversampling}. Thus, we expect and confirm experimentally that it benefits from much lower variance compared to the plug-in estimator, as illustrated by the confidence intervals in Figures \ref{fig:discrete_results}, \ref{fig:continuous_results} (plots are in log-log scale). For a more quantitative analysis, we plot the \textbf{empirical standard deviation} of Algorithm \ref{our_algorithm} on the Gaussian dataset on Figure \ref{fig:standard_deviation_gaussian_alg1}. It is worth noticing that it exhibits a much lower variance no matter how clusterable the underlying distribution is. However, proving this requires bounding the stability of the optimal quantization solution, for which no directly applicable results exist. 

\begin{figure}[h]
     \centering
     \includegraphics[width=1.65in]{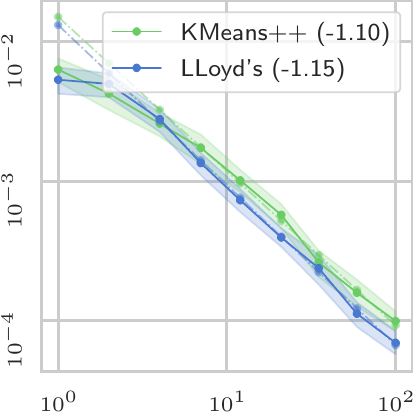}
     \caption{Mean relative error vs.\ $k$ on the Gaussian dataset 
     for Algorithm \ref{our_algorithm} (green), and the same algorithm succeeded by Lloyd's procedure (blue). The improvement of the latter is marginal and comes at the expense of few $O(nk)$ steps.
     }
     \label{fig:lloyd_refinement}
\end{figure}

\textbf{Lloyd's algorithm.}
$k$-means++ is often used as an initialization step for Lloyd's algorithm. The latter converges to a local minimizer of the quantization error, at the expense of few more passes through the data, for an overall complexity of $O(nki)$, where $i$ is the number of iterations. Theoretically, this algorithm makes the quantization error decay by $\log k$ at best. We verify experimentally that the improvement is marginal in Figure \ref{fig:lloyd_refinement}.

\subsection{Algorithm 2}
To test the performance of Algorithm \ref{our_algorithm_for_regularized_transport}, we compare it to do an approximate solver for entropy-regularized optimal transport, which is arguably the most popular occurence in machine learning applications. Specifically, for datasets $(\mu_n, \nu_n)$, we measure the CPU time to execute Algorithm \ref{our_algorithm_for_regularized_transport} 
with input $(\mu_n, \nu_n, \epsilon)$ and $\approxsolver(\mu_n, \nu_n, 3\epsilon)$, which are both guaranteed to output a $3\epsilon-$approximation of OT. Here, $\approxsolver$ is from \citep{DBLP:journals/corr/AltschulerWR17}, but any other approximate solver satisfying the same constraints on the input/output can be used. We display two types of plots: \emph{(i)} CPU time vs.\ precision $\epsilon$ and \emph{(ii)} estimated transport cost vs.\ precision $\epsilon$. The former demonstrates efficiency while the latter shows that the output is indeed at most $\epsilon$ away from the unregularized cost. 

\begin{figure}
\vspace*{-12pt}
\hspace{-.1in}
\begin{minipage}{3.34in}
    \centering
    \begin{tabular}{c c}
         $(t, \epsilon)$ & $(\mathrm{error}, \epsilon)$ \\
         \includegraphics[width=.47\textwidth]{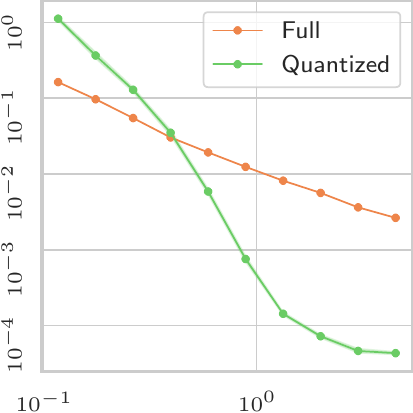} &  \includegraphics[width=.47\textwidth]{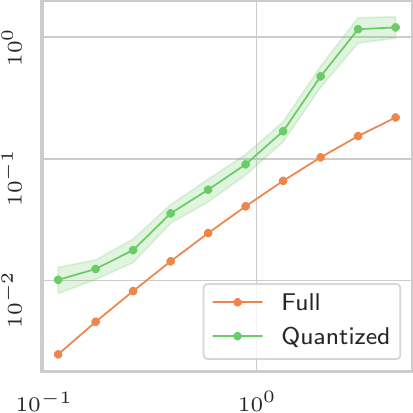} \\
         (a) Gaussian, $\sigma= 10^{-1}$ . & (b) Gaussian, $\sigma= 10^{-1}$.  \\
         \includegraphics[width=.47\textwidth]{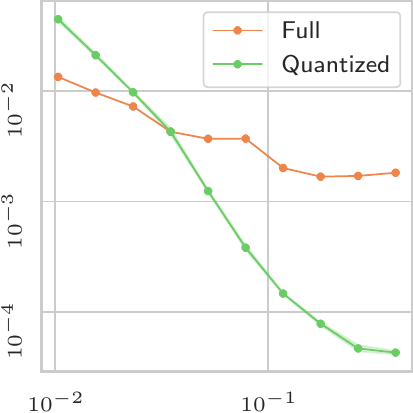} &  \includegraphics[width=.47\textwidth]{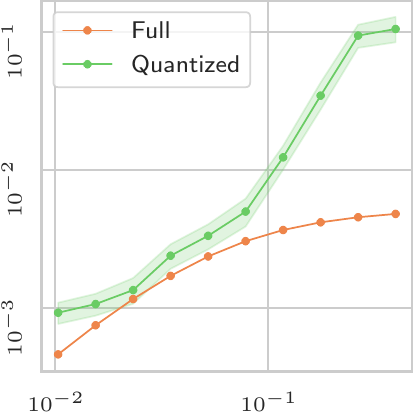} \\
         (c) Gaussian $\sigma=10^{-3}$ . & (d) Gaussian, $\sigma=10^{-3}$. \\
         \includegraphics[width=.47\textwidth]{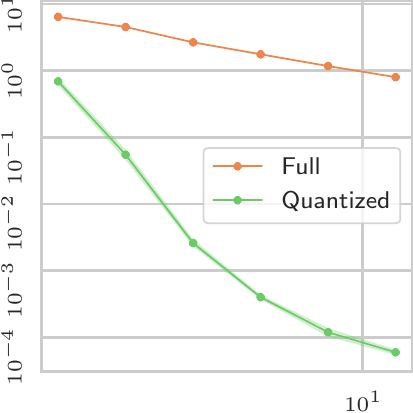} &    \includegraphics[width=.47\textwidth]{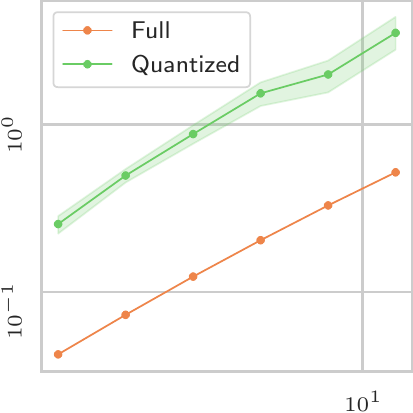} \\ 
         (e) Adult Dataset. & (f) Adult Dataset.
    \end{tabular}
\caption{Left: CPU time ($s$) vs.\ $\epsilon$, for Algorithm \ref{our_algorithm_for_regularized_transport} and $\approxsolver$ \citep{DBLP:journals/corr/AltschulerWR17}. Right: absolute error vs.\ $\epsilon$. The smallest precision for the range of $\epsilon$ is taken so that $\approxsolver$ requires $n_{\mathrm{max}}=10^4$ iterations. Algorithm \ref{our_algorithm_for_regularized_transport} consistently provides an approximate solution an order of magnitude faster than $\approxsolver$. }
\label{fig:regularized_experiments}
\end{minipage}
\vspace*{-12pt}
\end{figure}

\textbf{Results.} From the CPU time plots in fig.\ \ref{fig:regularized_experiments} (left column)
the speedup introduced by our algorithm is unmistakable. It only matches the performance of $\approxsolver$ for low values of $\epsilon$, when $\quantize$ simply outputs the whole dataset to have a small enough quantization error. That's why it is most useful for structured data, e.g.\ peaked distributions (fig.\ \ref{fig:regularized_experiments}.c)
or real-world datasets (fig. \ref{fig:regularized_experiments}.e)
The error vs.\ $\epsilon$ plots (right column) suggest that the bounds in \citep{DBLP:journals/corr/AltschulerWR17} are loose, since the error is often smaller than the guaranteed $\epsilon$. Quantization enables us to have maximum efficiency for bounded inaccuracy.

\section{Conclusion}
Our algorithm is designed with practicality in mind: at best---and in most of our experiments---we observe and expect reduced bias for fixed computational budget; at worst, it behaves like plug-in estimation. 
Our bounds explain the estimator's good behavior by relating $W_2$ to quantization error. 
Even when we fall back to the $\nicefrac{-1}{d}$ rate asymptotically, we have up to twice the decay rate in the finite sample case. 
Quantization is also efficient in aproximate OT solvers, as it can match their error with improved time/space complexity.

\begin{acknowledgements} 
The MIT Geometric Data Processing group acknowledges the generous support of Army Research Office grant W911NF2010168, of
Air Force Office of Scientific Research award FA9550-19-1-031, of National Science Foundation grant IIS-1838071, from the CSAIL Systems that Learn program, from the MIT–IBM Watson AI Laboratory, from the Toyota–CSAIL Joint Research Center, from a gift
from Adobe Systems, from an MIT.nano Immersion Lab/NCSOFT Gaming Program seed grant, and from the Skoltech–MIT Next Generation Program.
\end{acknowledgements}

\bibliography{beugnot_126}

\appendix

\newpage
\onecolumn
\title{Improving Approximations of Optimal Transport Distances with Quantization : \\ 
Supplementary Materials}
\maketitle

\setcounter{section}{0}
\section{Additional information on the experiments}
\label{section:appendix_experiments}
\subsection{Additional information on the setup}
\textbf{Implementation.}
We implemented the $k$-means++ subsampler using a combination of C++ and Python.
Although the complexity matches the bounds suggested in this article, our implementation was not designed with computational efficiency in mind:  later releases will provide faster implementation (i.e., lowering the constant in front of the $O(k^3 \log k)$). We used the transport LP solver of \cite{flamary2017pot}, but any other solver can be used in practice. 

\textbf{Datasets.}
In the \emph{gaussian} settings, we estimate the distance between a gaussian centered at ${\bf 0}_d$ and one centered at ${\bf 1}_d$, both with covariance $\tau \mathbb{I}_d$. $\tau$ is a parameter to see the influence of the clusterability on the performance of our estimator. The transport distance is known in closed form for gaussians, equal in this case to $\lVert {\bf 1}_d \rVert_2 = \sqrt{d}$. 

The \emph{fragmented-hypercube} 
is an example from \cite{forrow2018statistical}. It consists of the uniform distribution $\mu$ on $\csb{0, 1}^2$, which is pushed forward by the function $T(X) = X + 2 \; \mathrm{sign}(X) \odot {\bf 1}_2$. $T$ being the gradient of a convex function, the transport distance between $\mu$ and $T_\# \mu$ can be computed in closed form as $W_2 (\mu, T_\# \mu)= \sqrt{8}$. This example is extended to dimensions $d > 2$ by concatenating $\mu$ with $\mathcal{U}\csb{0, 1}^{d-2}$; while this does not change the transport cost, it adds statistical noise by mimicking a high-dimensional distribution with low-dimensional support. This definition enables a straightforward interpretation between the two quantities at play in Theorem~\ref{thm:optimal_bias_estimator}: in low dimension, our estimator has clear added value, but this efficiency is lost in the quantization error in higher-dimensional settings. 

The \emph{Sampled Mixtures} synthetic dataset is produced as follows. We sample ${\bf m}_1, {\bf m}_2 \sim \csb{0, 1}^{m \times d}$ and set $\Sigma = \tau \mathbb{I}_d$. Then, we draw $n_{\text{tot}}$ points from the mixture of gaussians $\cb{\mathcal{N}({\bf m}_{1, i}, \Sigma)}_{i=1}^m$ uniformly to  obtain a point cloud $X$. We do likewise for $Y \sim \p{\cb{\mathcal{N}({\bf m}_{2, i}, \Sigma)}_{i=1}^m}^{n_{\text{tot}}}$. We stress that once we sample these two point clouds, we do not sample again from the mixture of gaussians. The purpose of this experiment is to provide discrete point clouds in any dimension, with various shapes. 

The only preprocessing step applied to the \emph{Adult} dataset was centering and scaling.

\newcommand{\pointclouds}[2]{
 \scalebox{0.69}{
 \begin{tikzpicture}
        \begin{axis}[
        title=#1,
        enlargelimits=false,
        width=.44\linewidth,
        height=5cm,
        ticks=none,
        ]
        \addplot+ [only marks,mark size=0.6pt] table [x=x, y=y, col sep=comma] {#2_alpha.csv};
        \addplot+ [only marks,mark size=0.6pt] table [x=x, y=y, col sep=comma] {#2_beta.csv};
        \end{axis}
    \end{tikzpicture}
    }
}
\begin{figure}[h]
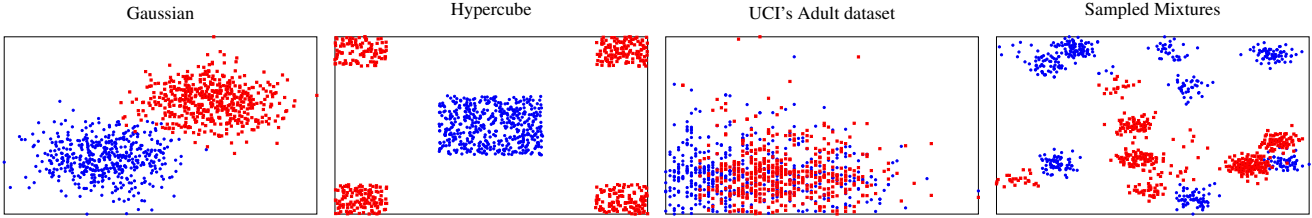

\centering
    \begin{tabular}{@{}c@{}c@{}c@{}c@{}}
        \pointclouds{Gaussian}{data/samples/gaussian_sample_500} &
        \pointclouds{Hypercube}{data/samples/hypercube_sample_500} &
        \pointclouds{UCI's Adult dataset}{data/samples/adult_sample_500} &
        \pointclouds{Sampled Mixtures}{data/samples/gaussian_mixture_sample_500_std3} 
    \end{tabular}
\caption{First two components of some distributions tested in our experiments.  Gaussian and Hypercube: continuous distributions (we display samples). UCI and Sampled Mixtures: discrete point clouds. \emph{UCI} is in $\RR^6$, and \emph{Sampled Mixtures} are random point clouds of $n_{\text{tot}}=10^4$ points.\vspace{-.15in}}
\label{tested_distributions}
\end{figure}

\textbf{CPU time simulations.} Wall clock time was measured using the CPU clock and included the whole pipeline: Sample $n$ points, subsample $k$ anchors, and run the linear program on the anchors. Each line in Figure~\ref{fig:cpu_time} has multiple points marked: each point corresponds to a different choice of $k$. For KMeans and AFK-MC\textsuperscript{2}, there are 9 values of $k$ evenly log-spaced from 1 to 100. For the na\"ive estimator, there are 15 values of $k$ ranging from 1 to 1000. An analogous procedure was used to generate the plots in Figure~\ref{fig:discrete_results}, with a different $x$-axis.

\subsection{Additional experiments}

\begin{figure}
     \centering
     \begin{subfigure}[b]{0.3\textwidth}
         \centering
         \includegraphics[width=\textwidth]{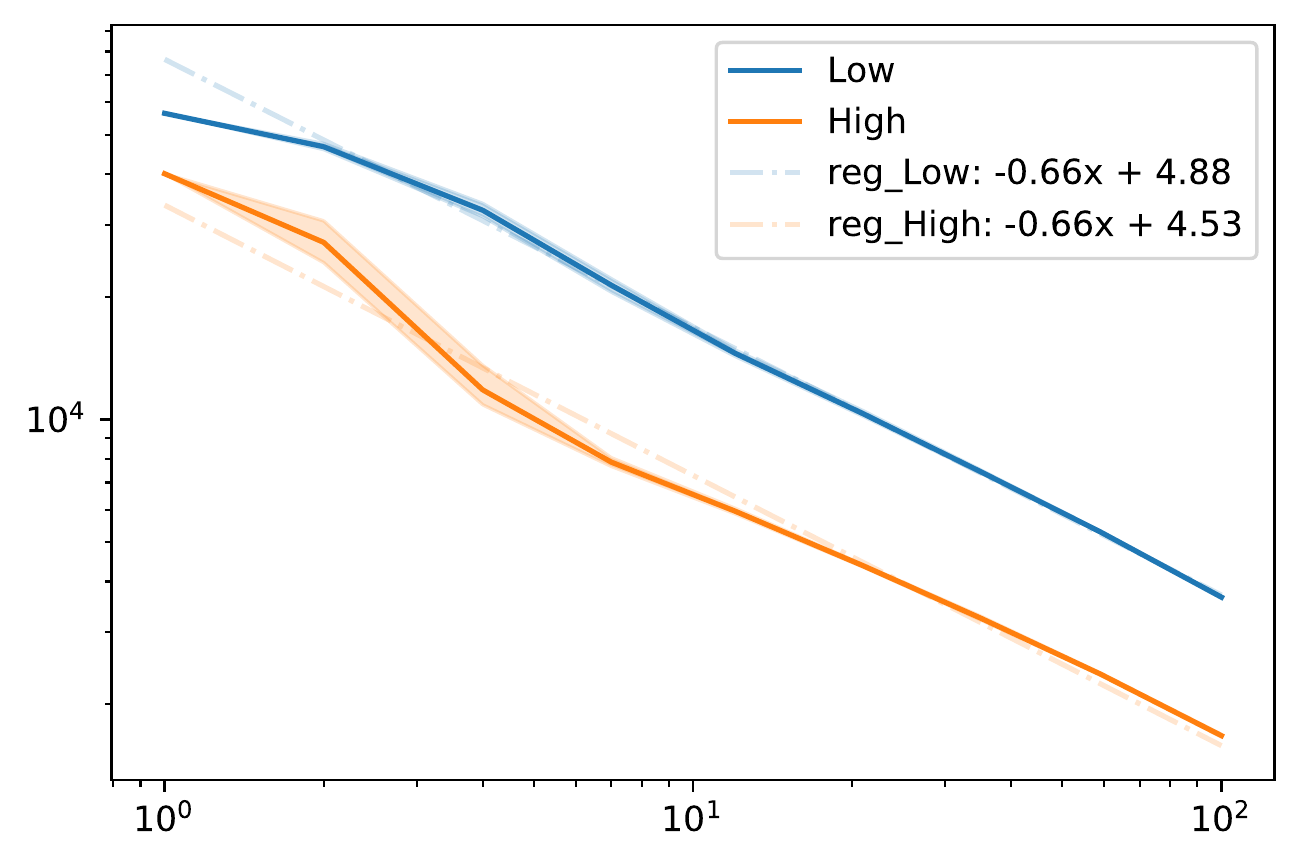}
         \caption{Adult}
     \end{subfigure}
     \hspace{12pt}
     \begin{subfigure}[b]{0.3\textwidth}
         \centering
         \includegraphics[width=\textwidth]{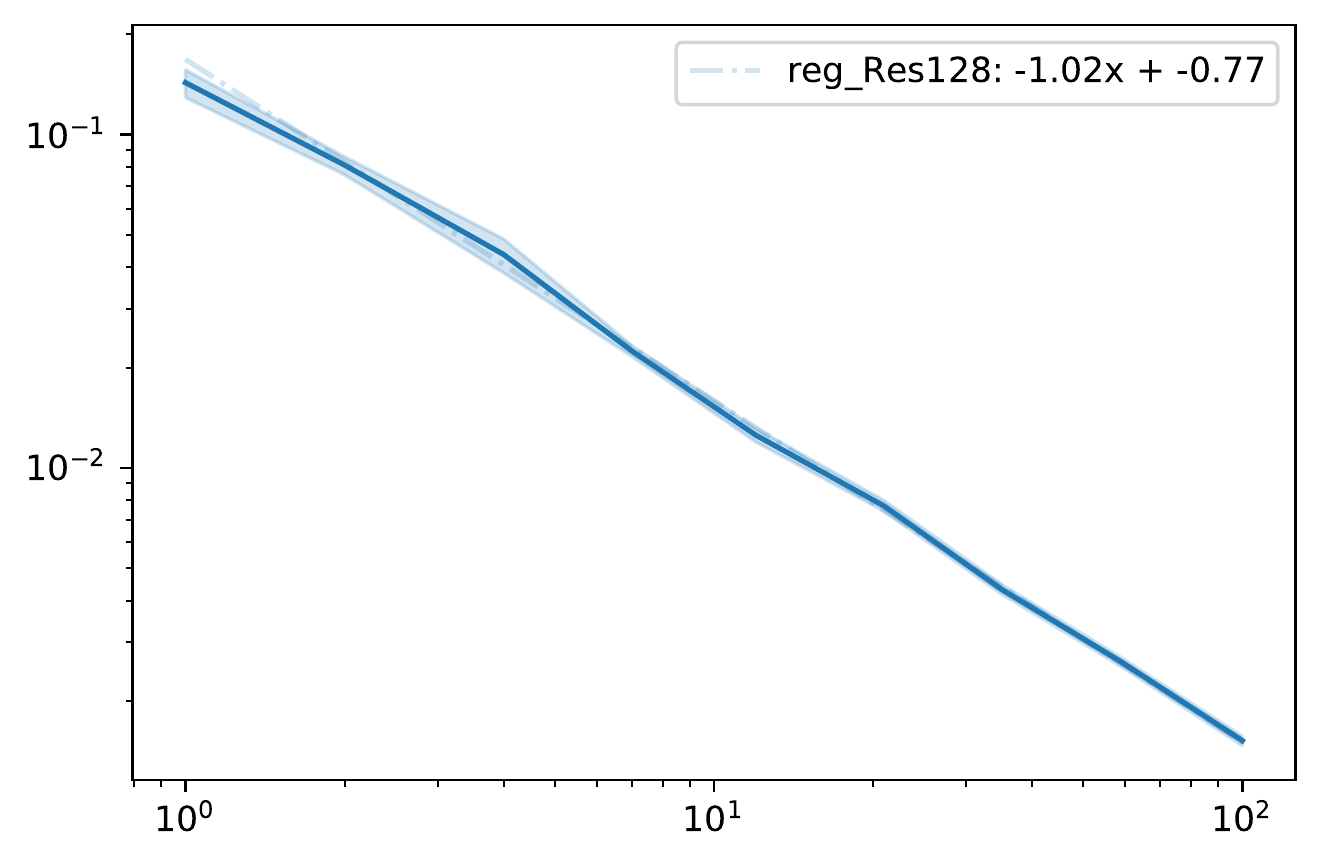}
         \caption{DOT, 128 Microscopy}
     \end{subfigure}
     \vspace{6pt}
     
     \begin{subfigure}[b]{0.3\textwidth}
         \centering
         \includegraphics[width=\textwidth]{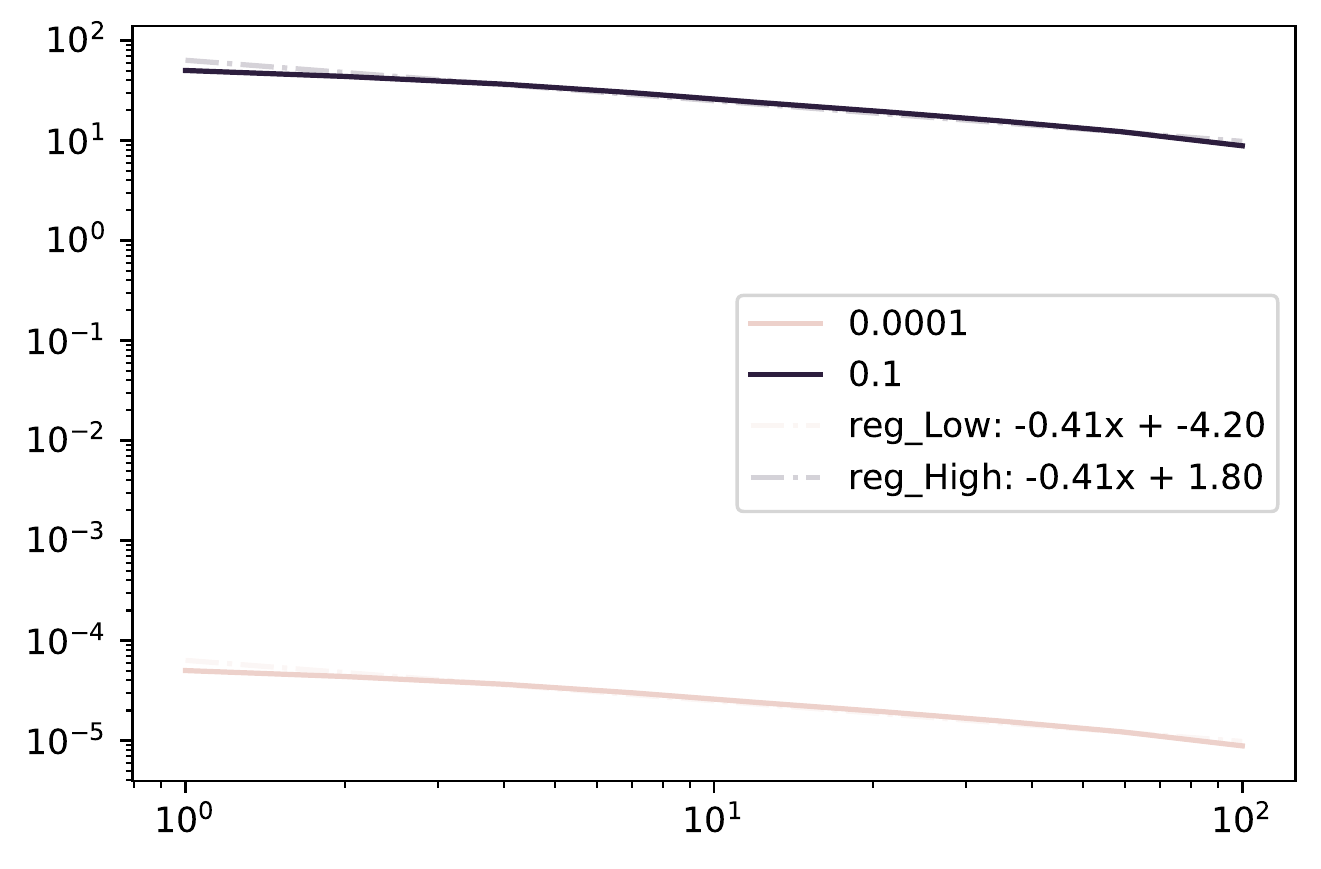}
         \caption{Gaussian in $d=5$}
     \end{subfigure}
     \begin{subfigure}[b]{0.3\textwidth}
         \centering
         \includegraphics[width=\textwidth]{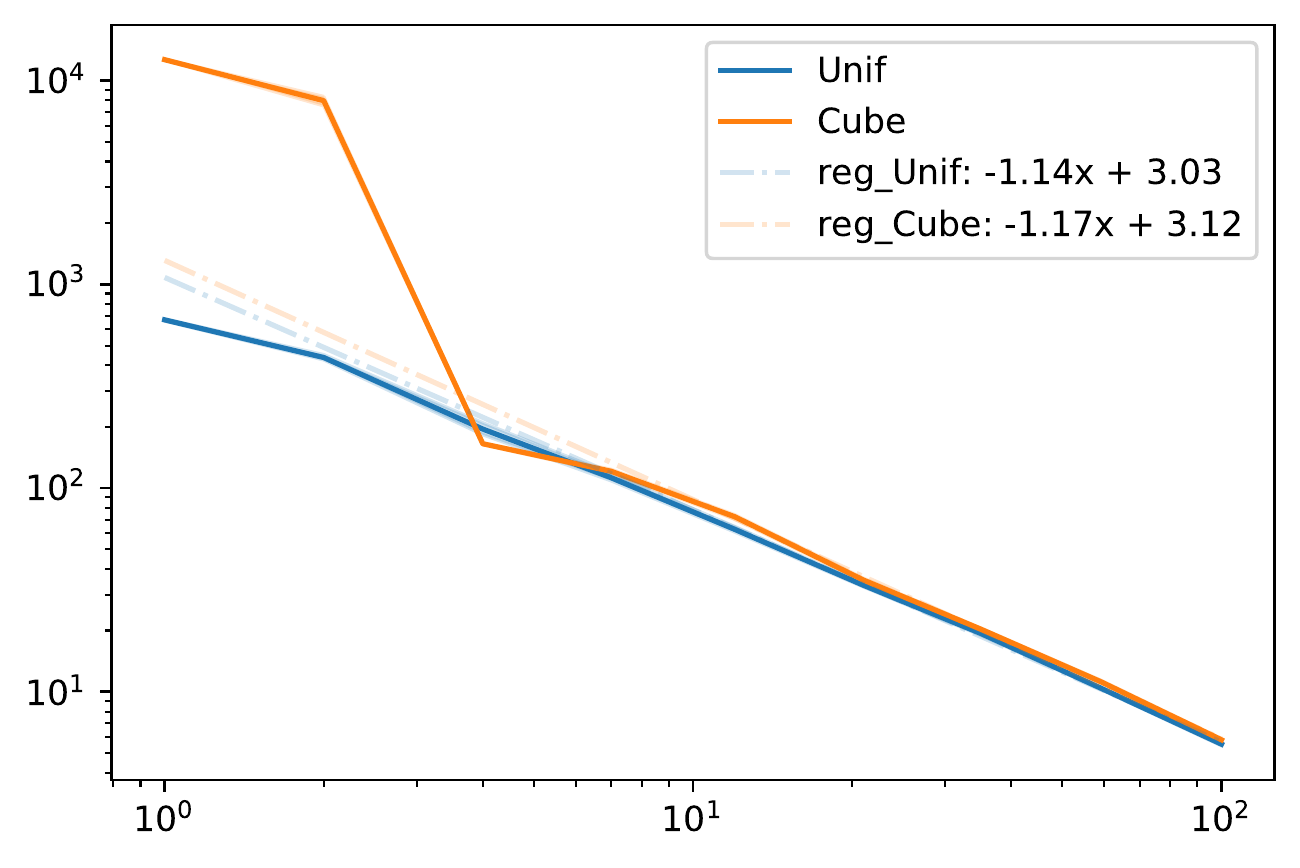}
         \caption{Hypercube in $d=2$}
     \end{subfigure}
     \begin{subfigure}[b]{0.3\textwidth}
         \centering
         \includegraphics[width=\textwidth]{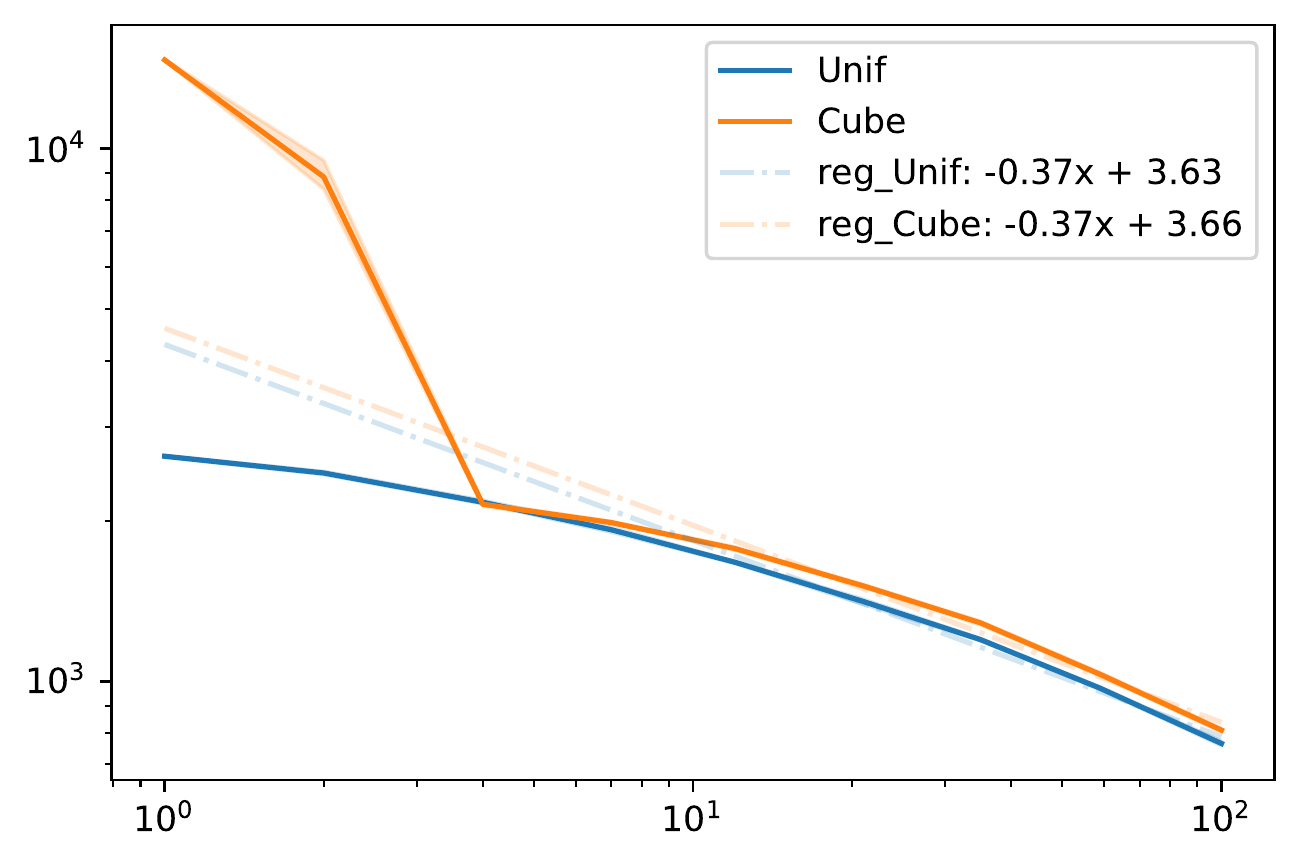}
         \caption{Hypercube in $d=8$}
     \end{subfigure}
        \caption{Quantization error vs. $k$. Same settings as described in the article. For the Adult dataset, both distributions are displayed (high and low income), and k-Means++ was ran 20 times for each. DOT plot's is the average over the 10 available distributions. The quantization error decays faster than the worst-case $1/d$ of the uniform distribution: $0.66 > 1/6$ for adult, and $1 > 1/2$ for DOT. $1000$ points were sampled from continuous distributions, and the output of $k$-Means was averaged over 10 times. For Gaussians, the slope is the same for both low and high variance, but low variance yields a much smaller quantization error, and thus better peformance of our method.}
        \label{fig:qerr}
\end{figure}

\paragraph{Quantized data assumption.} We provide plots in Figure \ref{fig:qerr} suggesting that the low quantization error  assumption made to quantify the sample complexity is verified for the real-world datasets we use. Remember that quantization doesn't improve the rate when the distribution is close to uniform, or when the scale at which we process the data is below the signal's scale. Such situation is unlikely to appear in real-world settings, where we want to compute distances between \textit{signals} rather than noise. 

\begin{figure}
\centering
    \begin{subfigure}[t]{1.6in}
         \includegraphics[width=\textwidth]{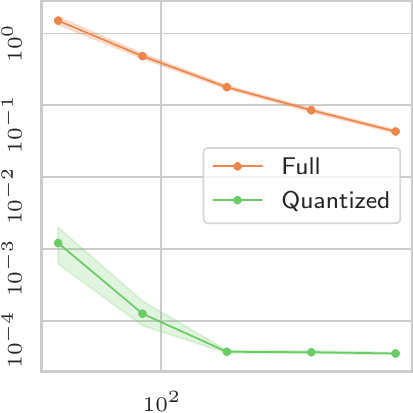}
         \caption{CPU times (\textit{s}) vs $\epsilon$}
    \end{subfigure}
    \hspace{-5pt}
    \begin{subfigure}[t]{1.6in}
         \includegraphics[width=\textwidth]{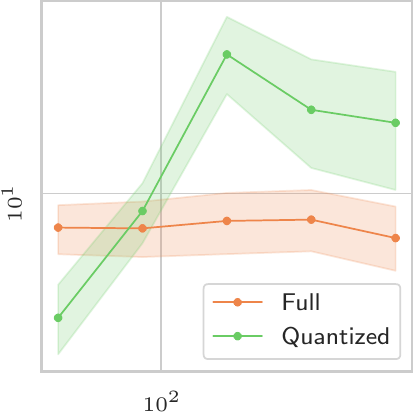}
         \caption{Absolute error vs. $\epsilon$}
    \end{subfigure}
\caption{Comparison between Algorithm \ref{our_algorithm_for_regularized_transport} and $\approxsolver$ on DOT, a benchmark of gray-scaled images for optimal transport solvers \cite{Schrieber_2017_DOT}. $\approxsolver$ comes from \cite{DBLP:journals/corr/AltschulerWR17}. 
For each $\epsilon$, values are averaged over the 45 pairwise estimation.}
\label{fig:regularized_dot}
\end{figure}

\paragraph{Algorithm \ref{our_algorithm_for_regularized_transport} on DOT.} Due to lack of space, we report in Figure \ref{fig:regularized_dot} the performance of Algorithm \ref{our_algorithm_for_regularized_transport} on subsampled images of DOT. Again, our estimator is a magnitude faster. The quantization step is well suited to the two dimensional support of images.

\vspace{-5em}

\end{document}